\documentclass[a4paper, 10pt, conference]{ieeeconf}      % Use this line for a4 paper

\IEEEoverridecommandlockouts                              % This command is only needed if 
\usepackage{amsmath,empheq}
\usepackage{graphicx}
\usepackage{float}
\usepackage[utf8]{inputenc}
\usepackage{hyperref}
\IEEEoverridecommandlockouts
\usepackage{bm}

\usepackage{caption}
\usepackage{subcaption}
\usepackage{dblfloatfix}
\usepackage{tikz}
\usetikzlibrary{arrows,automata,chains,calc,intersections}
\usetikzlibrary{backgrounds}

\begin{document}

\title{Task-Space Control Interface for SoftBank Humanoid Robots and its Human-Robot Interaction Applications}

\author{authors}

\author{Anastasia Bolotnikova$^{1,2}$, Pierre Gergondet$^{4}$, Arnaud Tanguy$^{3}$, Sébastien Courtois$^{1}$, Abderrahmane Kheddar$^{3,2,4}$
\thanks{$^1$SoftBank Robotics Europe, Paris, France}
\thanks{$^2$University of Montpellier--CNRS LIRMM, Interactive Digital Humans, Montpellier, France}
\thanks{$^3$CNRS-AIST Joint Robotics Laboratory, IRL 3218, Tsukuba, Japan}
\thanks{$^4$Beijing Advanced Innovation Center for Intelligent Robots and Systems, Beijing Institute of Technology, Beijing, China}}

\maketitle

\begin{abstract}
We present an open-source software interface, called {\tt mc\_naoqi}, that allows to perform whole-body task-space Quadratic Programming based control, implemented in {\tt mc\_rtc} framework, on the SoftBank Robotics Europe humanoid robots. We describe the control interface, associated robot description packages, robot modules and sample whole-body controllers. We demonstrate the use of these tools in simulation for a robot interacting with a human model. Finally, we showcase and discuss the use of the developed open-source tools for running the human-robot close contact interaction experiments with real human subjects inspired from assistance scenarios.
\end{abstract}

\IEEEpeerreviewmaketitle

\section{Introduction and Background}

Quadratic Programming (QP) task-space control has become a golden standard in humanoid robot control~\cite{escande2014ijrr, cisneros2019humanoids, bouyarmane2019tro}. One of the most powerful software control frameworks that implements QP task-space control, called {\tt mc\_rtc}\footnote{\url{https://jrl-umi3218.github.io/mc_rtc}}, is now publicly available. This framework provides developers with useful tools for writing complex controllers for either individual or multiple interacting robots to perform wide variety of experiments, e.g. in aircraft automation~\cite{kheddar2019ram} or in physically interacting with humans~\cite{otani2018icra, bolotnikova2020roman} that we technically explain and extend in this work.

In order to control any given complex robots with this framework, an interface must be developed to allow communication between the control framework and the robot's low-level controllers, sensors and devices. One example of such open-source interface is the {\tt mc\_openrtm}\footnote{\url{https://github.com/jrl-umi3218/mc_openrtm}} software that is used to control Humanoid Robotics Project robots from Kawada Robotics~\cite{hirukawa2004ras}. However, for different types of robots a specific interface, as well as robot description and robot module, need to be developed in order to adapt to particularities of robot brand and low-level control strategies, devices and onboard operating system (OS). The idea of task-space control is to lie exactly between low-level control and high-level planning with somewhat adjustable frontiers.

\begin{figure}[!t]
   \centering
    \includegraphics[width=\columnwidth]{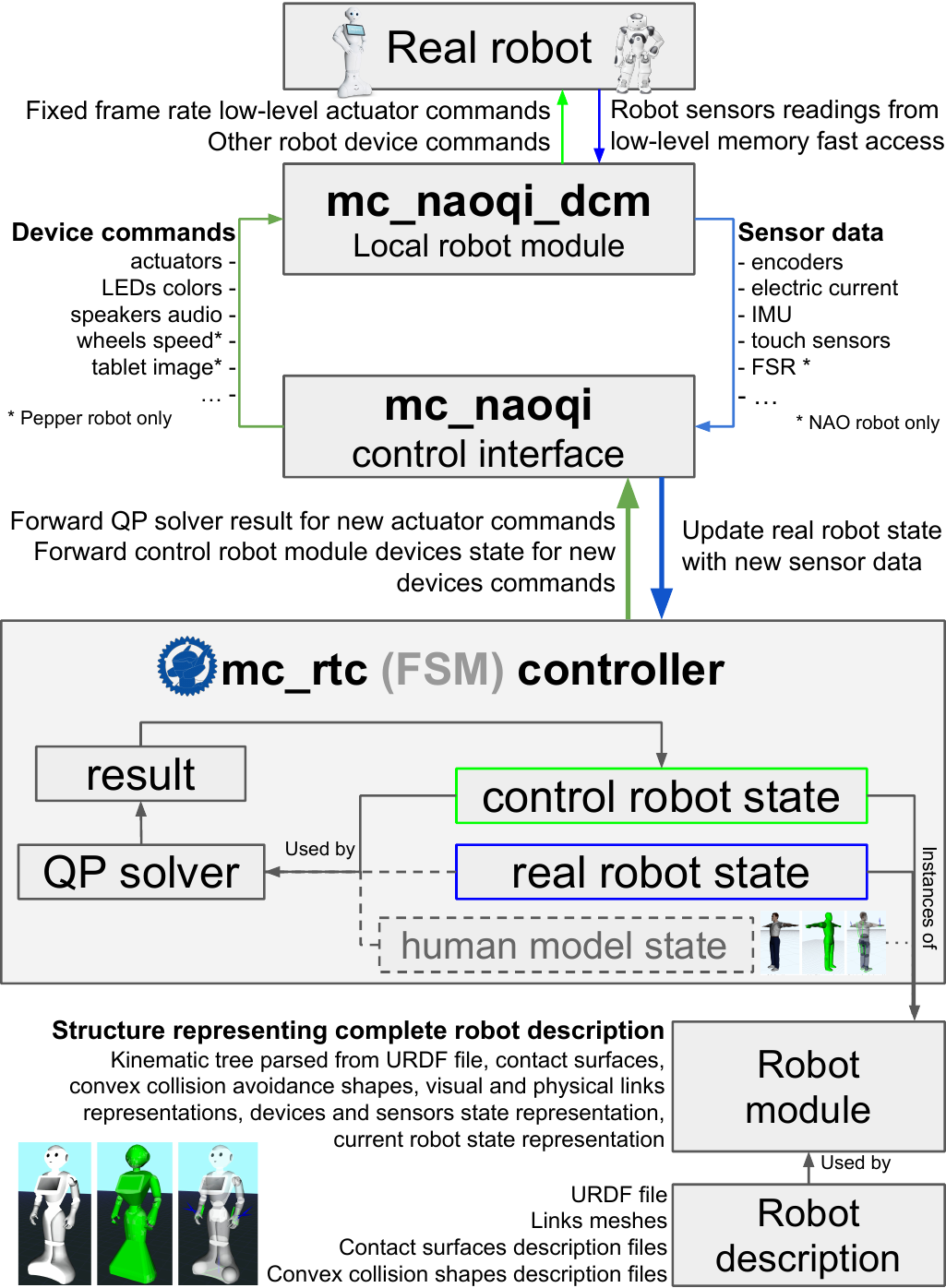}
    \caption{{\tt mc\_naoqi} interface enables communication between SBRE humanoid robots and {\tt mc\_rtc} control framework. It can be used to steer the robot behaviour in HRI experiments.}
    \label{fig:mcnaoqi}
\end{figure}

In this work, we present the open-source software interface, called {\tt mc\_naoqi}\footnote{\url{https://github.com/jrl-umi3218/mc_naoqi}}, that enables running the controllers implemented with {\tt mc\_rtc} framework on widely used SoftBank Robotics Europe (SBRE) humanoid robots~\cite{gouaillier2009icra, pandey2018ram}, that are running NAOqi OS and a customly developed local module {\tt mc\_naoqi\_dcm} (Sec.~\ref{sec:interface}).
We also provide robot modules and description packages that serve as a representation of SBRE robots in the {\tt mc\_rtc} framework (Sec.~\ref{sec:robotmodule}). Additionally, we release a basic sample Finite State Machine (FSM) {\tt mc\_rtc} controller for Pepper robot and its interaction with a human model in simulation (Sec.~\ref{sec:samplecontroller}). Finally, we present our current instance use of the developed tools for running the {\tt mc\_rtc} controllers developed for Human-Robot interaction (HRI) experiments with human subjects (Sec.~\ref{sec:experiments}). 

Fig.~\ref{fig:mcnaoqi} shows an overview of the developed tools and their interconnection, described in detail in the rest of the paper.

\section{Control Interface}
\label{sec:interface}

\begin{figure}[!htb]
   \centering
    \includegraphics[width=\columnwidth]{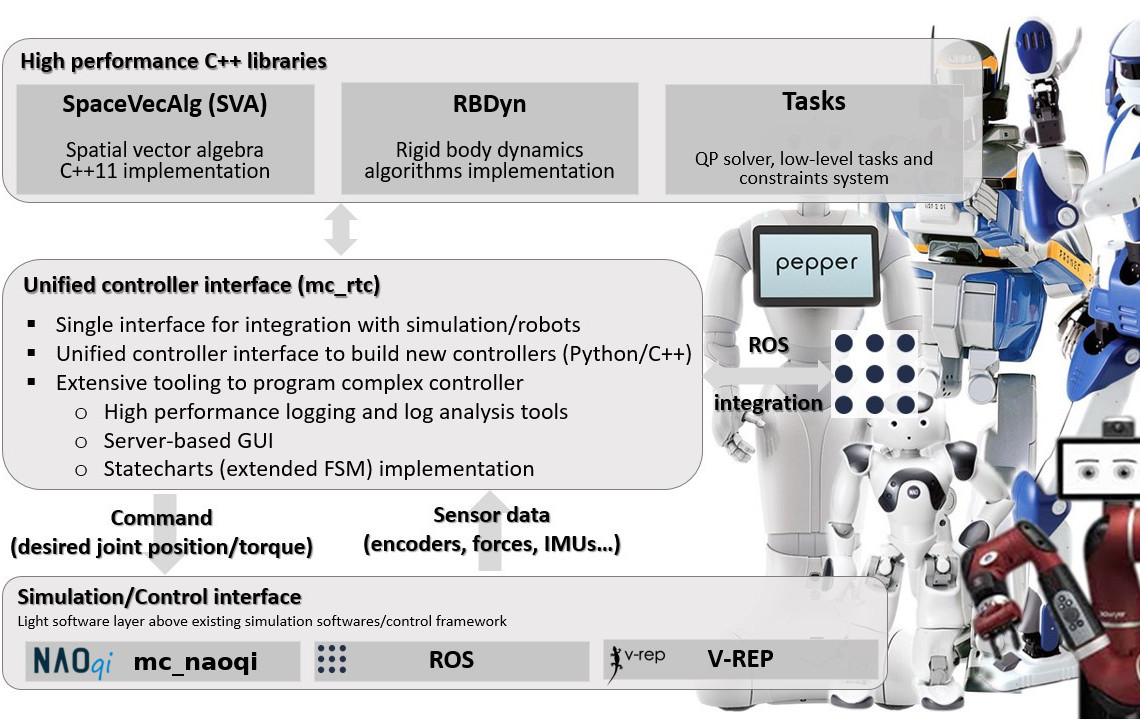}
    \caption{{\tt mc\_rtc} control framework architecture.}
    \label{fig:mcrtc}
\end{figure}

Fig.~\ref{fig:mcrtc} gives a schematic overview of the {\tt mc\_rtc} control framework and its connection to the simulation and control interfaces, such as {\tt mc\_naoqi}. Interested readers are invited to visit {\tt mc\_rtc} project website for more information, installation instructions and tutorials.

Fig.~\ref{fig:mcnaoqi} illustrates schematically the role of the {\tt mc\_naoqi} interface as a communication layer between {\tt mc\_rtc} control framework and NAOqi OS running onboard SBRE humanoid robots, such as NAO, Pepper or Romeo. In  the following subsections, we describe the entire system in detail.

\subsection{Forwarding device commands from controller to the robot}

The main role of the {\tt mc\_naoqi} interface is to forward the fixed frame rate control commands computed by the QP solver of {\tt mc\_rtc} whole-body QP controller to the onboard low-level robot actuators control. The decision variable of the {\tt mc\_rtc} QP controller is robot joint accelerations, which is integrated once to get desired velocity (e.g. for Pepper mobile base command), and then once more to get desired robot joint angles for joint actuator commands.

For social, user-friendly and interacting humanoid robots, such as Pepper or NAO, it is highly beneficial to endow {\tt mc\_rtc} controller with the functionality to also forward other device commands, such as sentence to play from the speakers, desired tablet screen image or eye led color, from the {\tt mc\_rtc} controller to the robot devices via {\tt mc\_naoqi} interface. This functionality allows {\tt mc\_rtc} framework users to develop controllers which can provide a richer interaction experience for HRI applications. For instance, when a certain {\tt mc\_rtc} controller FSM state is terminated, the robot can indicate this event (and that it is going to transition to the next FSM state) with a comprehensive verbal message, illustrative tablet image and/or a specific LED color.

The {\tt mc\_naoqi} interface is thus developed with rich HRI consideration in mind, and allows {\tt mc\_rtc} controller to forward commands to all the robot devices, not only the joint actuators. We describe how it is implemented on the robot module side in detail in Sec~\ref{sec:robotmodule}.

\subsection{Forwarding sensor data from the robot to the controller}

The {\tt mc\_naoqi} interface is also responsible for getting the most up-to-date sensor readings from a robot low-level memory in real-time and forwarding sensor measurements in a suitable form to the {\tt mc\_rtc} controller real robot state representation as a feedback. This way, the task-space QP controller keeps track of the real robot state and can use it to perform closed-loop QP control computations.

Besides common sensor readings, such as encoder values, force sensors or Inertial measurement unit (IMU) measurements, {\tt mc\_naoqi} interface also allows to forward from the robot to the {\tt mc\_rtc} controller such sensor readings as electric motor current and touch sensor readings (from tactile or bumper sensors). We describe how custom robot sensors are implemented on the robot module side in detail in Sec~\ref{sec:robotmodule}.

For HRI applications, the touch sensor readings are especially beneficial to be forwarded to the controller, as they allow to detect when a human touches the robot. This signal can then be used inside the {\tt mc\_rtc} FSM controller, for instance, to trigger an appropriate reaction of the robot to the touch or to trigger a transition to a specific FSM state of the controller. We demonstrate this use-case in our sample HRI experiment, presented in detail in Sec.~\ref{sec:experiments}.

\subsection{Local low-level robot module} 

A customized local robot low-level module, called {\tt mc\_naoqi\_dcm}\footnote{\url{https://github.com/jrl-umi3218/mc_naoqi_dcm}}, is cross-compiled for NAOqi OS and is set to run onboard the robot to read sensor values and set device commands synchronized with a robot control loop via Device Communication Manager (DCM) every 12~ms, fastest update rate currently available for SBRE robots.

When a user connects to the robot via {\tt mc\_naoqi} interface, and commands to turn on robot motors, a preprocess function is connected to DCM loop, to start setting the actuator commands at a fixed time rate, synchronized with the other DCM operations. After the user commands to turn off robot motors via {\tt mc\_naoqi} interface, the preprocess actuator command update function is disconnected from the DCM loop. This way, the user can safely use other control applications, such as \emph{Choregraphe}, with the robot right after using {\tt mc\_naoqi} interface, even though {\tt mc\_naoqi\_dcm} module is still active on the robot.

In order to prevent the default high-level robot behaviours to interfere with the commands forwarded to the robot via {\tt mc\_naoqi} from the {\tt mc\_rtc} controller, the default robot safety reflexes are disabled when robot motors are turned on via {\tt mc\_naoqi}. Once the motors are turned off via {\tt mc\_naoqi}, the default robot safety reflexes are re-enabled. This has to be taken into account by an {\tt mc\_rtc} controller designer, to ensure that the developed controller is safe to be run on the real robot, through extensive testing in simulation.

A fast access to the low-level robot memory is initialized  when {\tt mc\_naoqi\_dcm} starts to run on the robot. This allows to read a predefined set of sensor values from robot memory in the fastest way.

\section{Robot representation in {\tt mc\_rtc} framework}
\label{sec:robotmodule}

\subsection{Robot description packages}

\begin{figure} [!htb]
\begin{subfigure}{0.5\textwidth}
  \centering
  \includegraphics[width=\linewidth]{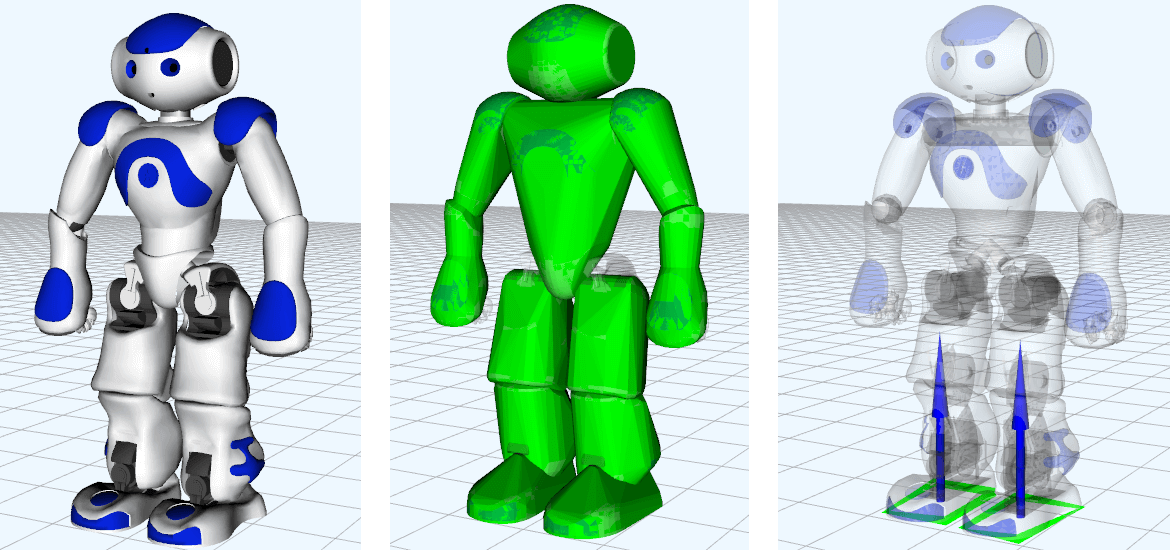}
\end{subfigure}

\vspace{0.15cm}

\begin{subfigure}{0.5\textwidth}
  \centering
  \includegraphics[width=\linewidth]{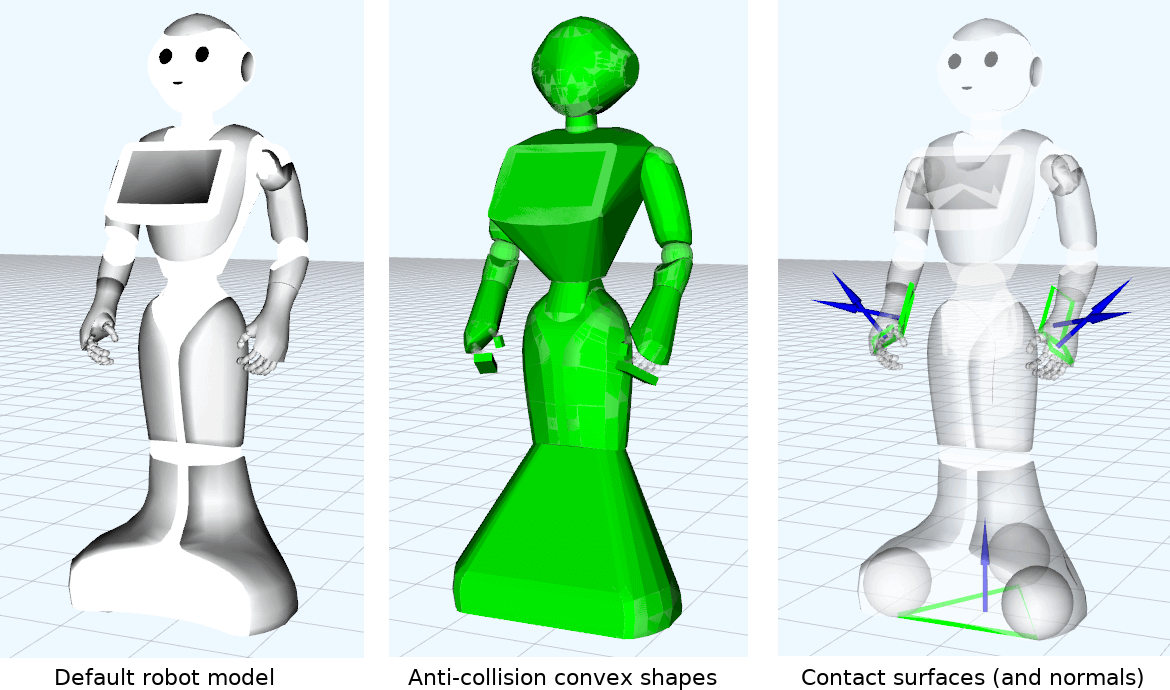}
\end{subfigure}
\caption{NAO and Pepper robot descriptions for the robot representation in the robot module of the {\tt mc\_rtc} framework.}
\label{fig:description}
\end{figure}

To control any robot with {\tt mc\_rtc} framework, a basic description of this robot needs to be provided. Such robot description includes robot kinematic tree, dynamic properties of the links, description of robot contact surfaces (i.e. covers), and convex (eventually strictly convex~\cite{escande2014tro}) anti-collision shapes (these shapes can be automatically generated from existing robot links graphic representation). Examples of these robot description elements for SBRE humanoids are shown in Fig.~\ref{fig:description}. With this work, we release these robot description projects, implemented as Robot Operating System (ROS) packages, for NAO\footnote{\scriptsize{\url{https://github.com/jrl-umi3218/nao_description}}} and Pepper\footnote{\scriptsize{\url{https://github.com/jrl-umi3218/pepper_description}}} robots.

\subsection{Robot {\tt mc\_rtc} modules}

As illustrated in Fig.~\ref{fig:mcnaoqi}, the robot description packages are used by a \emph{robot module} software layer to create a structure that provides a complete description of the robot: kinematic tree parsed from URDF files, visual and physical representations of robot links, surfaces attached to the robot bodies, sensors and other devices, strictly convex hulls and primitive shapes for collision avoidance, etc. The instance of this structure is used by the {\tt mc\_rtc} framework, as control robot state representation, to formulate the QP control problem (objectives and constraints), that is then passed to the solver to compute the next desired robot state. 

We make the robot modules for NAO\footnote{\scriptsize{\url{https://github.com/jrl-umi3218/mc_nao}}} and Pepper\footnote{\scriptsize{\url{https://github.com/jrl-umi3218/mc_pepper}}} publicly available with this work. For fast prototyping and experiments, the robot description package and module can easily be augmented with any new robot hardware elements, e.g. new onboard camera, which we demonstrate in Sec.~\ref{sec:experiments}.

The Pepper robot module exploits a recently introduced new {\tt mc\_rtc} framework feature - generic robot devices. This feature enables developers to implement any kind of robot custom device representation as part of the robot module, which can then be used in the {\tt mc\_rtc} controller. Currently implemented devices in Pepper robot module are:
\begin{itemize}
\item \emph{loud speaker}: to forward speech commands to the robot directly from the {\tt mc\_rtc} controller via {\tt mc\_naoqi};
\item \emph{visual display tablet}: to set robot tablet image from the {\tt mc\_rtc} controller via {\tt mc\_naoqi} interface;
\item \emph{touch sensor}: to forward tactile sensor readings from the robot to the {\tt mc\_rtc} controller via {\tt mc\_naoqi}
\end{itemize}

Wheeled mobile base of Pepper is modeled as a floating base, constrained to move on a plane (i.e. the room ground), with limits imposed to the mobile base body maximum speed and acceleration according to physical hardware limitations.

In {\tt mc\_rtc} framework, two main hierarchies for the robot control are the following:
\begin{itemize}
\item \emph{Tasks}: that are objectives `describing' what the robot should do at the best it can;
\item \emph{Constraints}: that are limits under which the tasks are to be performed, i.e. what the robot should always fulfill strictly.
\end{itemize}
Many tasks and constraints are already implemented as robot-independent templates in {\tt mc\_rtc}. For instance \emph{PostureTask}, \emph{CoMTask}, \emph{EndEffectorTask}, \emph{KinematicsConstraint},  \emph{ContactConstraint}, etc. However, in some cases it might be desirable to design and implement new custom tasks or constraints, not yet implemented in {\tt mc\_rtc}. Such new tasks and constraints might be specific to a robot, use-case or research topic. 
In the Pepper robot module {\tt mc\_pepper}, we provide an example of the implementation of a custom \emph{CoMRelativeBodyTask} QP control objective. This task allows to specify the desired Pepper center of mass (CoM) target relative to the robot mobile base frame (as opposed to world frame in {\tt mc\_rtc} \emph{CoMTask}). The implementation of this Pepper specific objective was necessary to allow a controller to simultaneously compute a new mobile base position and keep the CoM objective as part of QP computations.
An example of a Pepper specific QP constraint implementation is a custom \emph{BoundedAccelerationConstr} constraint, included in the Pepper robot module. This constraint allows to impose acceleration bounds for Pepper mobile base link.

An example of how these custom Pepper robot specific tasks and constraints are loaded and used in a sample {\tt mc\_rtc} controller can be seen in \emph{PepperFSMController} open-source project, which we describe in detail in Sec.~\ref{sec:samplecontroller}.
In an analogous way, many other novel tasks and constraints can easily be implemented and tested.

\section{Sample Pepper {\tt mc\_rtc} FSM controller}
\label{sec:samplecontroller}

\subsection{Individual robot controller}

To facilitate the process of writing a new {\tt mc\_rtc} controller, especially for new potential users of the framework, we provide a basic sample \emph{PepperFSMController}\footnote{\scriptsize{\url{https://github.com/jrl-umi3218/pepper-fsm-controller}}}. It can be used as a starting point for new controller development or as an example of how similar projects should be implemented.

\begin{figure}[!htb]
   \centering
    \includegraphics[width=\columnwidth]{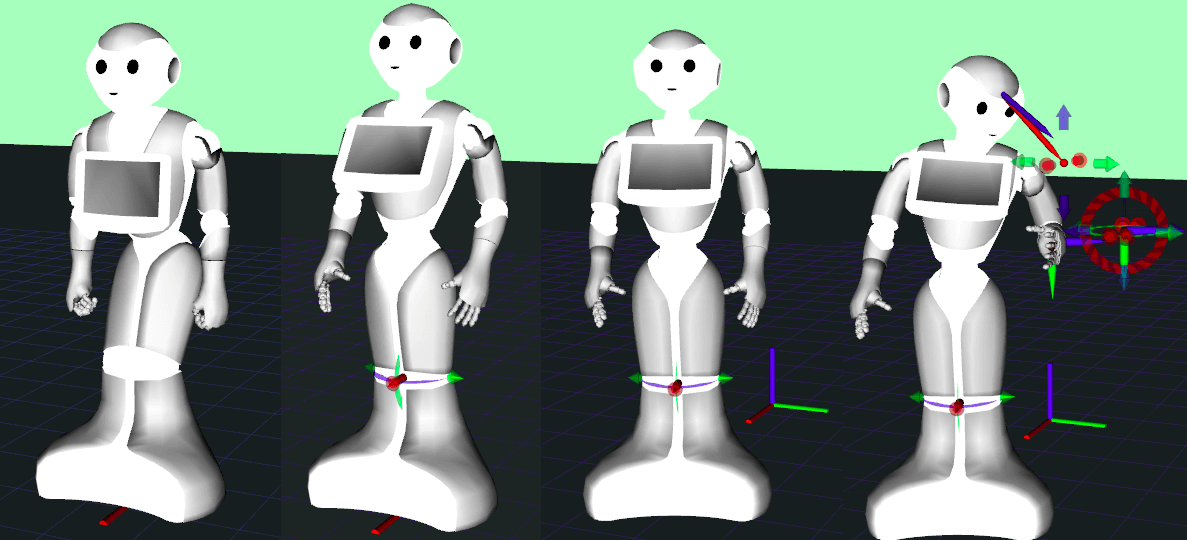}
    \caption{RViz scenes of sample FSM Pepper controller states.}
    \label{fig:rviz}
\end{figure}

The sample controller includes Pepper as the main controller robot, associated default posture task, kinematics and dynamics constraints, self-collision avoidance constraints, robot-ground contact constraint. The controller also includes implementation of few sample FSM states that allow to control robot posture, mobile base, camera orientation and right and left hand end-effector positions, either through predesined setpoints or interactively through RViz~\cite{kam2015ts} (Fig.~\ref{fig:rviz}).

\subsection{Controller for HRI including a human model}

\begin{figure}[!htb]
   \centering
    \includegraphics[width=\columnwidth]{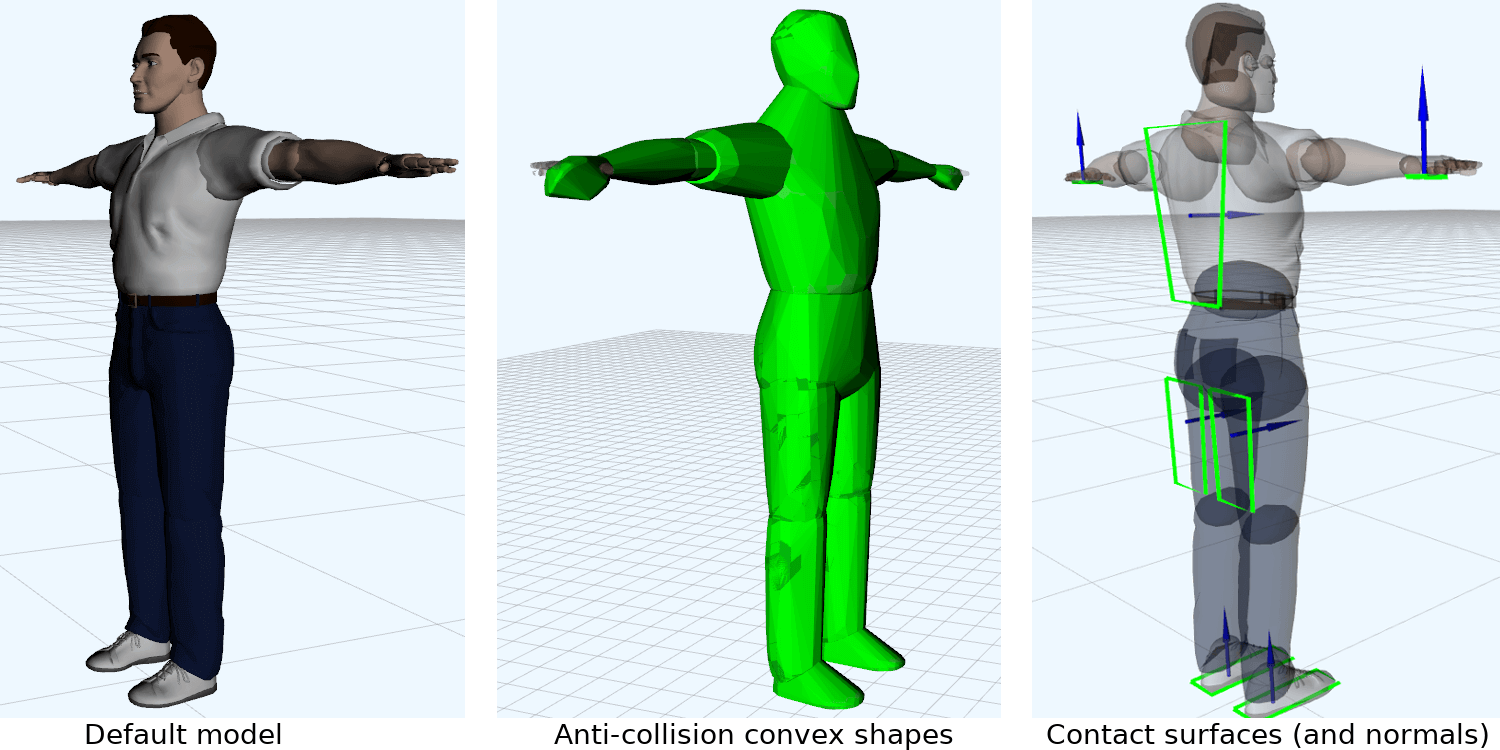}
    \caption{Simulation human model used in {\tt mc\_rtc} framework.}
    \label{fig:human}
\end{figure}

A direct extension of the {\tt master} branch of the sample controller project, is a branch called {\tt topic/withHumanModel}.
%\footnote{\url{https://github.com/jrl-umi3218/pepper-fsm-controller/tree/topic/withHumanModel}}.
On this branch, a multi-robot QP (MQP) control feature of the {\tt mc\_rtc} framework is exploited by adding a human model and its state as part of the {\tt mc\_rtc} controller (recall Fig.~\ref{fig:mcnaoqi}). 
A human model is integrated into {\tt mc\_rtc} exactly the same way that any other robot model, by providing a description ROS package\footnote{\scriptsize{\url{https://github.com/jrl-umi3218/human_description}}} and implementing a corresponding robot module\footnote{\scriptsize{\url{https://github.com/jrl-umi3218/mc_human}}}. Fig.~\ref{fig:human} shows the human model description used in our projects.

This MQP controller can be used to develop and simulate a wide variety of HRI scenarios including pHRI ones, e.g. using human model and robot contact surfaces description to define contact tasks and constraints. 

\begin{figure}[!htb]
   \centering
    \includegraphics[width=\columnwidth]{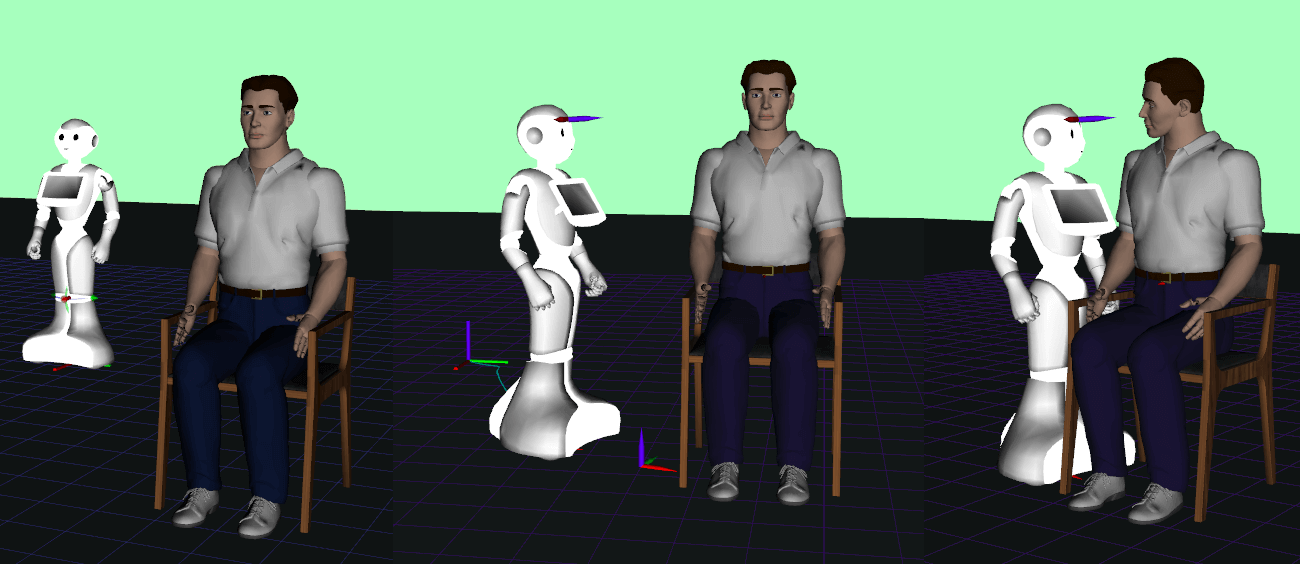}
    \caption{Simulated \emph{NavigateToHuman} state of a sample MQP {\tt mc\_rtc} FSM controller with a human model included.}
    \label{fig:navigatetohuman}
\end{figure}

In the open-source sample MQP controller project, we provide an example of an HRI controller state called \emph{NavigateToHuman}. In this state, a Position Based Visual Servoing (PBVS) task of {\tt mc\_rtc} framework is used to control Pepper robot mobile base to navigate in closed-loop to a set-point defined w.r.t human model torso frame (it can be any other human model frame), while using simulation data as a virtual visual feedback signal. At the same time, Image Based Visual Servoing (IBVS) task is used to control robot camera orientation to look at the human head. Fig.~\ref{fig:navigatetohuman} illustrates the simulation of the \emph{NavigateToHuman} FSM state at the start, middle and the end of this state.

The sample controller project can also efficiently serve as a starting point for developing controllers for various real HRI applications, which we showcase in Sec.~\ref{sec:experiments}

\section{Experiments with real human subject}
\label{sec:experiments}

Recently, the Covid-19 pandemic has put attention into robotics, which can help to deal with the problematics of human virus transmission, by transferring some risky labours and non-added value tasks of care-givers to robotic systems. Examples come from different companies, like disinfecting ultraviolet (UV) rovers and ground drones for hospitals, or even beaming videos to
connect patients to their relatives. There are more examples in other fields, like flagging patients with suspected pneumonia during their hospital admission, or using robots and AI in logistics to transport daily groceries.

\begin{figure*}
\begin{subfigure}{.24\textwidth}
  \centering
  \includegraphics[width=\linewidth]{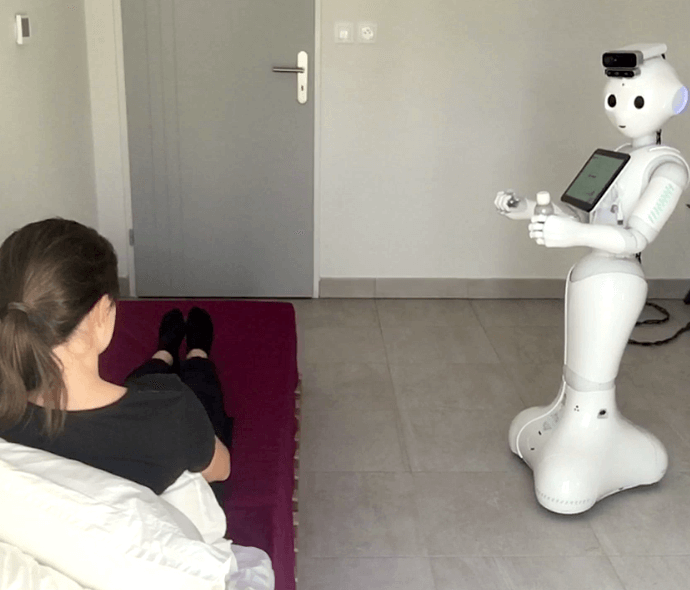}
\end{subfigure} \hfill
\begin{subfigure}{.24\textwidth}
  \centering
  \includegraphics[width=\linewidth]{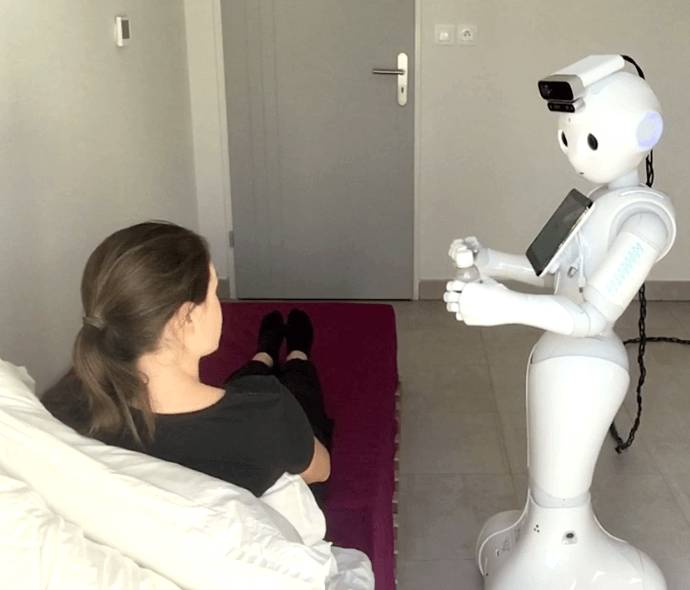}
\end{subfigure} \hfill
\begin{subfigure}{.24\textwidth}
  \centering
  \includegraphics[width=\linewidth]{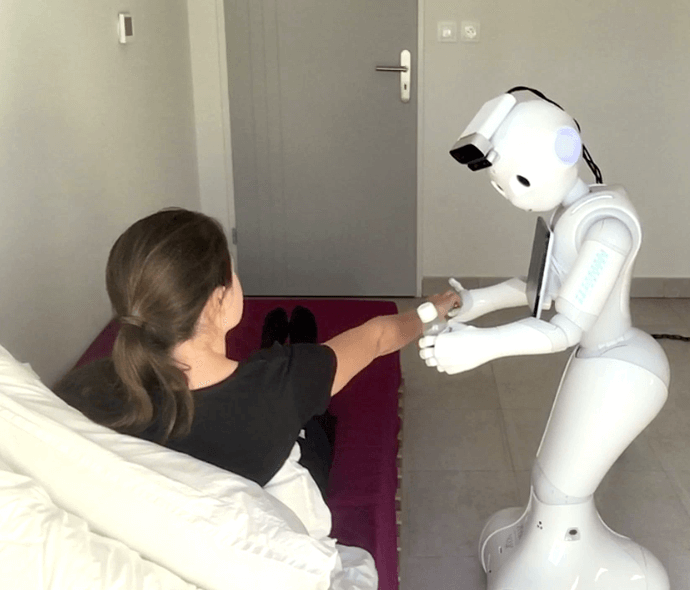}
\end{subfigure} \hfill
\begin{subfigure}{.24\textwidth}
  \centering
  \includegraphics[width=\linewidth]{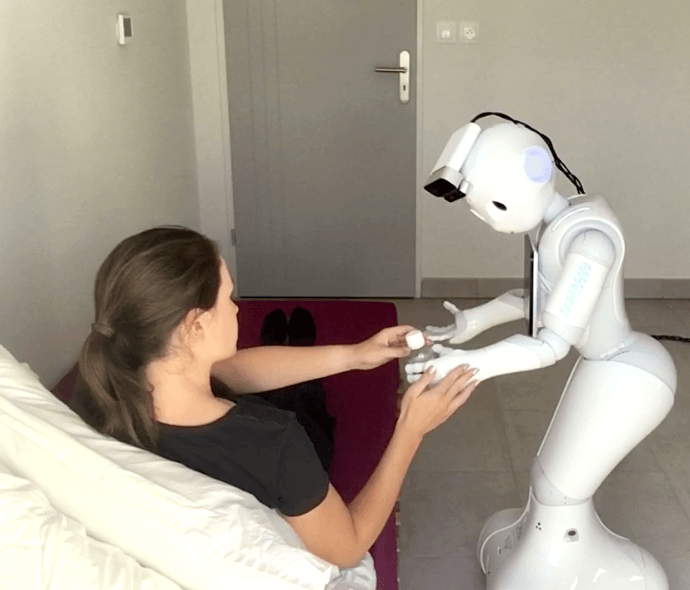}
\end{subfigure}

\vspace{0.1cm}

\begin{subfigure}{.24\textwidth}
  \centering
  \includegraphics[width=\linewidth]{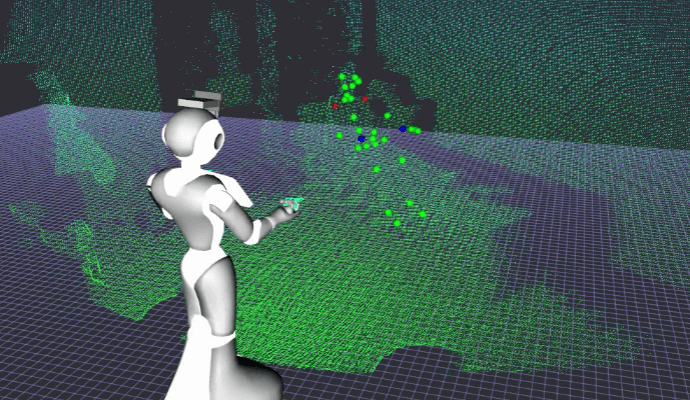}
  \caption{Navigation to the human}
  \label{fig:sfig1}
\end{subfigure} \hfill
\begin{subfigure}{.24\textwidth}
  \centering
  \includegraphics[width=\linewidth]{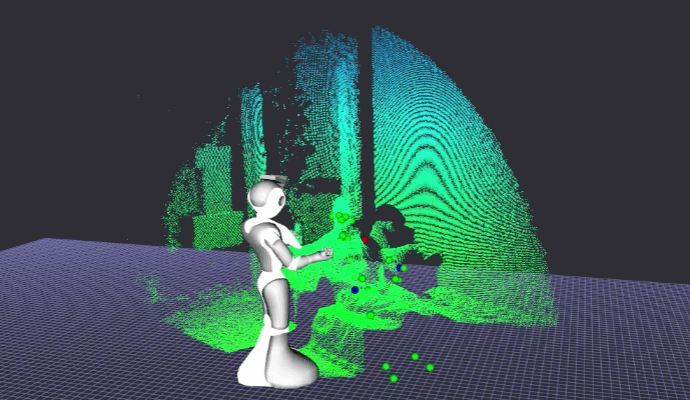}
  \caption{Verbal communication}
  \label{fig:sfig2}
\end{subfigure} \hfill
\begin{subfigure}{.24\textwidth}
  \centering
  \includegraphics[width=\linewidth]{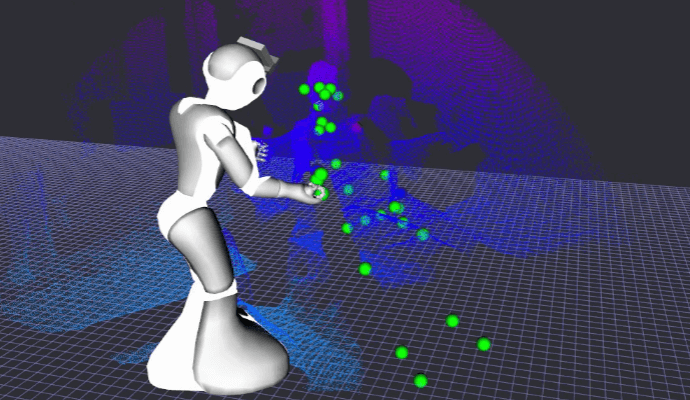}
  \caption{Human takes the pills}
  \label{fig:sfig3}
\end{subfigure} \hfill
\begin{subfigure}{.24\textwidth}
  \centering
  \includegraphics[width=\linewidth]{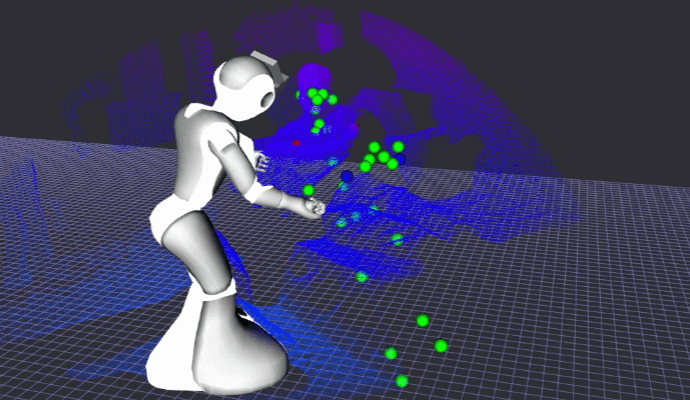}
  \caption{Human takes the bottle}
  \label{fig:sfig4}
\end{subfigure}
\caption{Experiment screenshots (top), perceived robot state, Azure Kinect point cloud and human body markers (bottom).}
\label{fig:entireexp}
\end{figure*}

In our recent publication, we have already demonstrated how the developed tools, presented in the current work, can be used to enable Pepper robot to perform autonomous initiation of human physical assistance~\cite{bolotnikova2020roman}. We also made the controller code\footnote{\scriptsize{\url{https://github.com/anastasiabolotnikova/autonomous_phri_init}}} publicly available to serve as a software reference to our work and as an advanced FSM {\tt mc\_rtc} HRI controller example. This controller shows how our developed software components allow to easily create complex controllers for rich, intuitive and efficient HRI. This includes closed-loop navigation toward human, verbal, visual and body language communication, and physical interaction with a human for an assistance process initiation.

In the present work, we develop another HRI application --inspired from the Covid-19 outbreak, where Pepper robot behaviour is regulated, with the help of the {\tt mc\_naoqi} and the developed robot modules, to autonomously deliver medication to a human lying on the bed. In this scenario a bed height conforms that of Pepper for the given tasks. 

The aim here is also to demonstrate how the controllers for new HRI scenarios can be easily prepared using our developed tools and re-using states and elements of the controllers written for other HRI scenarios. As such, for instance, the navigation toward a human state could be re-used from our previous work in~\cite{bolotnikova2020roman}, for this new scenario with almost no modifications, despite the person lying in the bed instead of sitting on a chair. Other parts of this new controller FSM also needed only slight adjustments w.r.t our existing work (states present in the other HRI controller) to create a controller for this new HRI application.

\begin{figure}[!htb]
   \centering
    \includegraphics[width=\columnwidth]{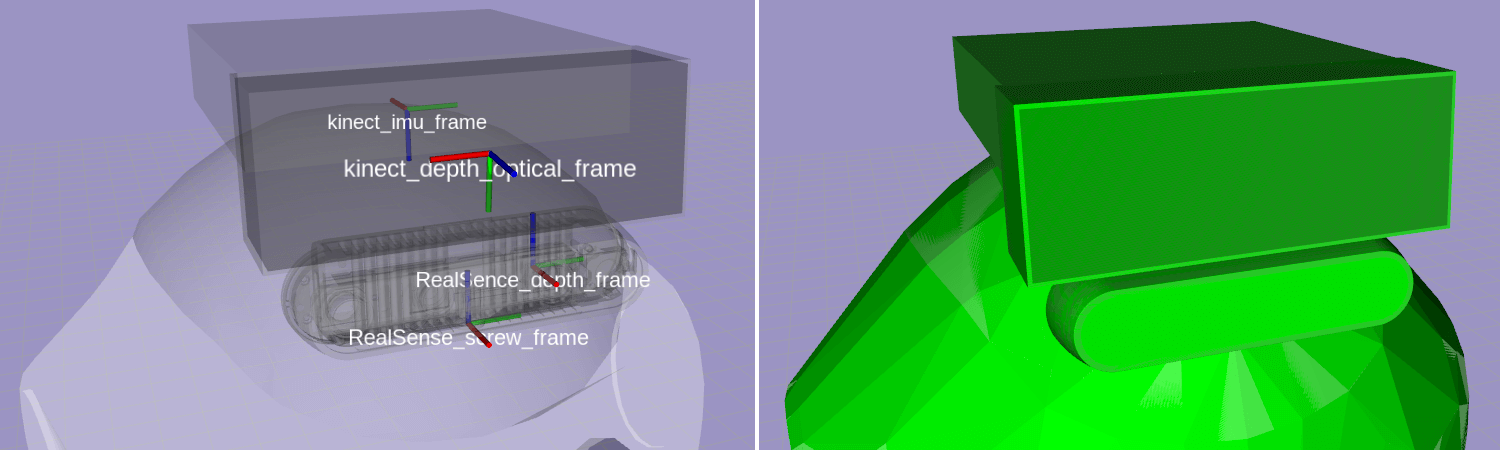}
    \caption{Extra hardware included in the prototype Pepper robot}
    \label{fig:extrahw}
\end{figure}

Fig.~\ref{fig:entireexp} shows excerpts screenshots from the experiment video. In the bottom row images, the scenes are visualized in RViz. Note, that additional hardware, namely Azure Kinect, which is used for human state feedback, and RealSense camera, which is included in the robot prototype, but not used in this experiment, are easily included in the robot description (Fig.~\ref{fig:extrahw}) and processed by the robot module, and therefore are also included in the QP problem formulation (e.g. for more accurate robot center of mass computation). Full experiment can be seen in the accompanying video. 

Once the robot reaches a position nearby the person, it communicates verbally its intention to pass the medicine (Fig.~\ref{fig:sfig2}). Then it proceeds to open its right gripper for the person to take the pills (Fig.~\ref{fig:sfig3}). The passing of the bottle with liquid is  arranged with the help of the robot module feature, described in Sec.~\ref{sec:robotmodule}, that allows to forward robot tactile sensor data to the {\tt mc\_rtc} controller, which then triggers robot left hand gripper to open slightly to allow the person to get out the bottle more easily (Fig.~\ref{fig:sfig4}).

\begin{figure}[!htb]
   \centering
    \includegraphics[width=\columnwidth]{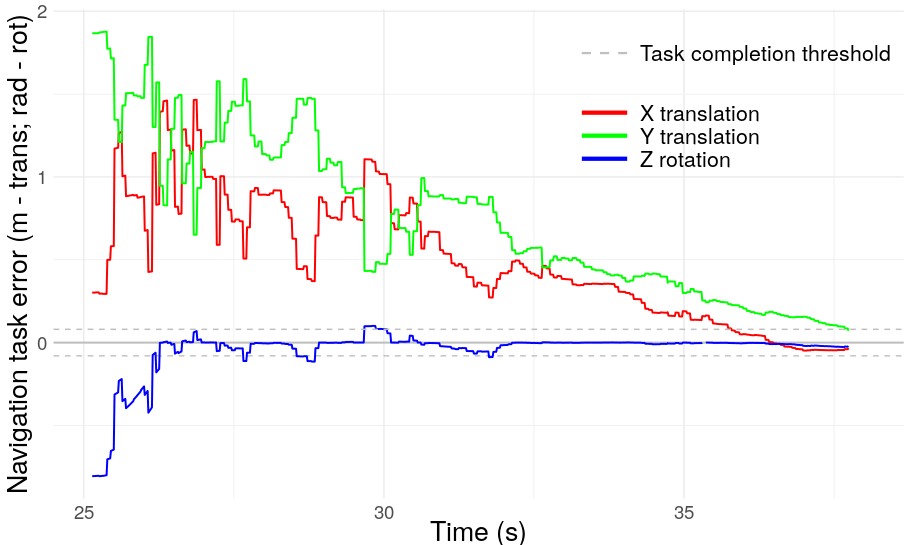}
    \caption{Evolution of Position Based Visual Servoing closed-loop Pepper mobile base navigation to human task errors}
    \label{fig:navigation_result}
\end{figure}

Fig.~\ref{fig:navigation_result} shows the progress of the closed-loop PBVS task, that uses Azure Kinect body tracking for feedback to make the robot navigate to the person (Fig.~\ref{fig:sfig1}). Task errors eventually converge near zero, although in a not very smooth way. This is due to the low quality of human detection (that prohibits constant usage in a continuous closed-loop way), and the low detection frame rate and latency issues (that limits the speed in reaching the person).
%, although the current speed is acceptable to meet safety and human acceptability constraints
%Note, however, that the modular architecture of our software tools allows to easily plug any other human detection tool in place of Azure Kinect, shall a better performing human detection software become available. 
Indeed, the existing human motion trackers are not robust enough for robotic usage. It is important to underline some shortcomings that prohibit current spreading of HRI in real practice:
\begin{itemize}
\item reliable human posture and body detection algorithms (even those advanced) often fail in robotics because of the usage perspective. In assistive robotics scenarios, it is the robot which moves toward a human that can be static (i.e. not moving much), e.g. lying in a home or hospital bed or sitting on a chair. Most of the human pose detection algorithms are trained with rather static camera and moving human and not with a moving camera and static (or moving) human. We have witnessed considerable effects on the robustness of the human pose acquisition by a moving robot with all the algorithms we tried. As a consequence, it is difficult to use them continuously in closed-loop control.
\item lack of ground-truth: most of the human pose detection matches well in an augmented-reality, their image or video counterpart, but no one provides measurement ground-truth concerning the pose results that are returned. That is to say, it is difficult to assess the precision of the 6D pose returned by most algorithms with ground-truth 6D measurements. In robotics, it is important to know precisely and in real-time the exact joint and floating base values of the human posture and configuration, namely when it comes to contact a person or to manipulate a given limb of a person~\cite{bolotnikova2020roman}.
\item last but not least, as it is the most critical issue: in close-contact interaction with humans (i.e. when the robot reaches a person), tracking may be lost even when a wide fish-eye camera is used. This is even more critical when physical interaction causes an obstructed view of the person. This means that human pose estimation approaches in the framework of human-robot close-contact interactions have to be deeply reviewed.
\end{itemize}

This being said, the approach we adopt is rather modular enough to live with such shortcoming without recalling into question the controller. Any improvement in human pose estimation w.r.t the performances mentioned previously will straightforwardly result in better robustness and performance of the controller, because of its conceptual simplicity.

\section{Conclusion}

For robotics to gain more insight, trust and meet the challenge of future human societal stakes, such as home daily assistance for frail or elderly persons, or to be efficiently used in future outbreaks, research efforts shall be paired with important integration development ones. Any advances made in critical human-robot interaction technological bricks such as human perception, artificial intelligence chatbots, 5G, advanced SLAM… should be integrated in readily available and sustained task-space control frameworks. This is the very reason of this work: we do not only aim at sharing the open-source code for the robotic community in general and the HRI in particular, but also sharing experiences that led to existing developments which are made to be further used, improved and sustained. Our methodology in terms of control can be seen now under a different philosophy. 

In computer science, algorithms, simple instructions such as if-then-else, while-for loops, variables... together with basic well-agreed routines and functions constitute the basic bricks of any modern algorithm that solve increasingly more complex problems. Our robot control framework should be seen under this spectrum: we shall provide elementary controllers and controller ``routines’’, templates that form the common ``instructions'', and functions of more complex controllers that solve more and more complicated tasks. In this paper, we thoroughly exemplify what {\tt mr\_rtc} may bring under a unified framework in terms of task specification, embedding straightforwardly the constraints... how a multi-(sensory,objectives,robots) task-space controller can be used to build sustainable controllers, and show that what we propose is consistent, as all these tasks are defined exactly the same way for any robot or multi-robot system. We hope that users of SBRE robots could assess, enrich and use our developments in even more challenging scenarios.

In future work, we are aiming to use this framework to conduct \emph{in-situ} assistance scenarios in a retirement house with few real patients. This needs to be complemented with considerable software engineering efforts.    

\section*{Acknowledgment}
We thank Sébastien Barthelemy and Sébastien Dalibard from SBRE for providing useful information for this work.

\bibliographystyle{ieeetr}
\bibliography{SII2021}

\end{document}